\documentclass[11pt]{article}

\usepackage[T1]{fontenc}
\usepackage[utf8]{inputenc}
\usepackage[margin=1in]{geometry}
\usepackage{microtype}
\usepackage{graphicx}
\usepackage{booktabs}
\usepackage{array}
\usepackage{tabularx}
\usepackage{multirow}
\usepackage{amsmath,amssymb}
\usepackage{caption}
\usepackage{subcaption}
\usepackage{enumitem}
\usepackage{float}
\usepackage{xcolor}
\usepackage{hyperref}
\usepackage{titlesec}
\usepackage{fancyhdr}
\usepackage{mathptmx}

\hypersetup{
  colorlinks=true,
  linkcolor=black,
  citecolor=black,
  urlcolor=black,
  pdftitle={Coupled Control, Structured Memory, and Verifiable Action in Agentic AI (SCRAT: Stochastic Control with Retrieval and Auditable Trajectories)}
}

\newcolumntype{Y}{>{\raggedright\arraybackslash}X}
\newcolumntype{L}[1]{>{\raggedright\arraybackslash}p{#1}}
\titleformat{\section}{\Large\bfseries\color{black}}{\thesection.}{0.5em}{}
\titleformat{\subsection}{\large\bfseries\color{black}}{\thesubsection.}{0.5em}{}

\title{\vspace*{-1.2em}\Large\bfseries Coupled Control, Structured Memory, and Verifiable Action in Agentic AI (SCRAT - Stochastic Control with Retrieval and Auditable Trajectories): \\
\normalsize A Comparative Perspective from Squirrel Locomotion and Scatter-Hoarding \\[-0.15em]}

\author{Maximiliano Armesto \and Christophe Kolb\\[0.35em]
\small Taller Technologies\\
\small \texttt{maximiliano.armesto@tallertechnologies.com} \quad \texttt{christophe.kolb@tallertechnologies.com}}
\date{April 2, 2026}

\begin{document}
\maketitle
\vspace{-1.2em}

\begin{abstract}
Agentic AI is increasingly judged not by fluent output alone but by whether it can act, remember, and verify under partial observability, delay, and strategic observation. Existing research often studies these demands separately: robotics emphasizes control, retrieval systems emphasize memory, and alignment or assurance work emphasizes checking and oversight. This article argues that squirrel ecology offers an unusually sharp comparative case because arboreal locomotion, scatter-hoarding, and audience-sensitive caching couple all three demands in one organism. We synthesize species-specific evidence from fox, eastern gray, and, in one field-memory comparison, red squirrels, and we impose an explicit evidence ladder throughout: empirical observation, minimal computational inference, and stronger AI design conjecture. We then introduce a minimal hierarchical partially observed control model with explicit latent environmental dynamics, structured episodic memory, observer-belief state, option-level actions, and delayed verifier signals. Framed this way, the squirrel literature motivates three primary hypotheses for agentic AI: (H1) fast local feedback plus predictive compensation should improve robustness under hidden dynamics shifts; (H2) memory organized for future control rather than archival recall should improve delayed retrieval under cue conflict and load; and (H3) verifiers and observer models should sit inside the action-memory loop, reducing silent failure and information leakage while remaining vulnerable to checker misspecification. A fourth, downstream conjecture is that role-differentiated proposer/executor/checker/adversary systems may reduce correlated error when information access and verification burden differ, although this claim is not established by the squirrel evidence itself. We therefore present the paper primarily as a comparative perspective and benchmark agenda rather than as a new theorem or comprehensive empirical benchmark study. The main contribution is a disciplined program of falsifiable claims about the coupling of control, memory, and verifiable action.
\end{abstract}


\begin{center}
\includegraphics[width=\textwidth]{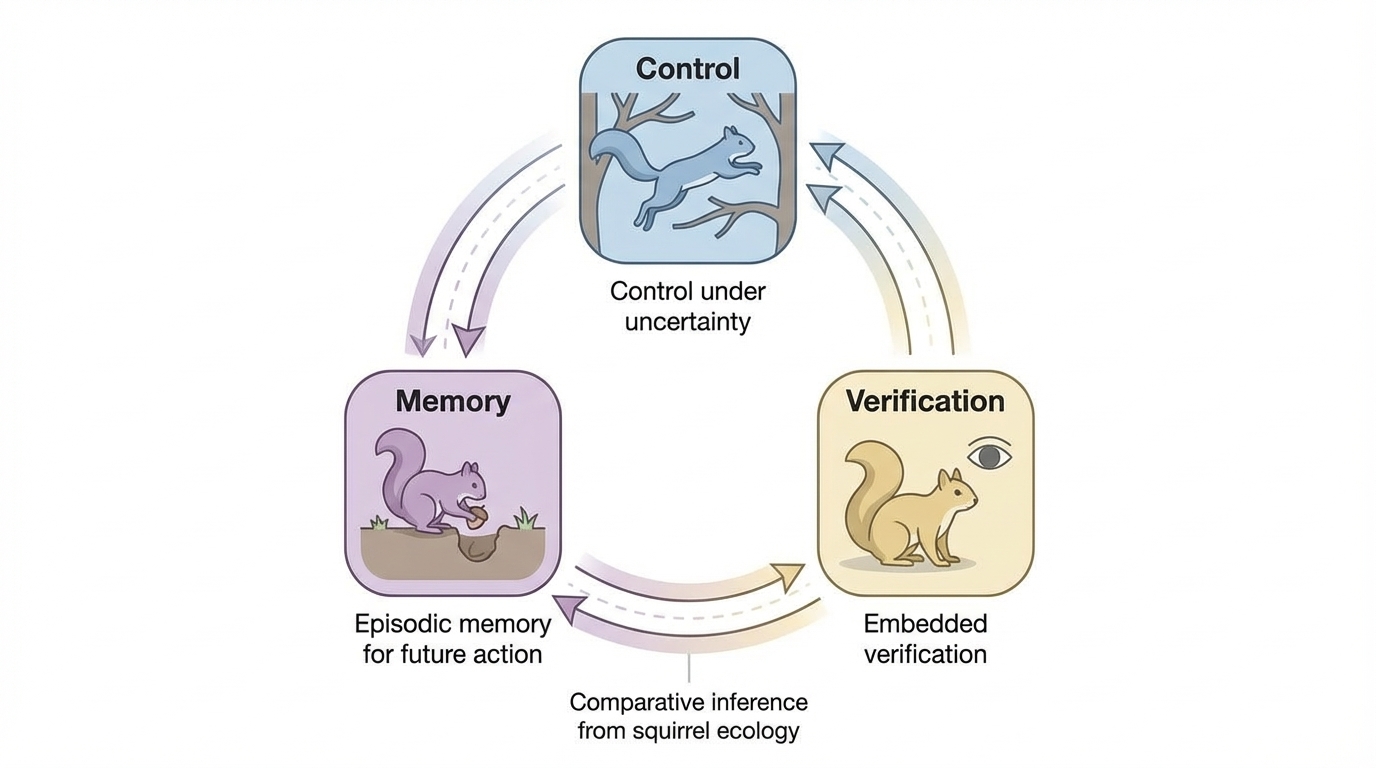}
\captionof{figure}{Conceptual overview of the coupled control–memory–verification problem studied in this paper. Squirrel behavior illustrates three tightly linked components: control under uncertainty, episodic memory for future action, and embedded verification under observation. The resulting loop motivates the hypothesis that agentic systems must integrate fast feedback control, structured memory, and in-loop verification to achieve robust performance under partial observability, delay, and strategic observation.}
\label{fig:abstract-overview}
\end{center}

\section{Introduction}

Many contemporary failures in agentic systems are not failures of local fluency. They are failures of coupling. A system writes a plausible plan but cannot recover after a local disturbance; retrieves a relevant document but acts on the wrong latent state; or satisfies a superficial checker while violating the user's operative objective. These pathologies point to a precursor problem that is narrower than alignment in general but central to deployment: how should an intelligent system be organized when rapid control, long-horizon memory, and verification must interact under uncertainty?

Squirrel ecology is a useful comparative case because it forces that coupling. Fox squirrels adapt leaps under uncertain support mechanics and improve on unfamiliar compliant branches [1]. Eastern gray squirrels recover their own caches with substantial spatial precision after delays and under naturalistic landmark conditions [2], [3]. Fox squirrels sometimes organize caches in a way consistent with chunking-like mnemonic structure and scale assessment and cache effort to value and scarcity [4], [5]. Gray squirrels also alter caching behavior when conspecifics can observe them [6], while broader reviews emphasize that food-hoarding decisions are shaped by social competition and theft risk [7]. The same ecological loop therefore links rapid control, delayed retrieval, valuation, and strategic observability.

The value of this comparison is not substrate similarity, and it is certainly not moral analogy. It is a shared computational problem family: hidden-state inference, delayed outcomes, memory-dependent action, and observer-sensitive policy. Those demands are familiar to robotics, reinforcement learning, retrieval-augmented systems, and AI assurance. What squirrels contribute is a natural case in which the demands are not optional extras studied in isolation, but ordinary components of one organism's daily competence.

This article is a comparative perspective and research agenda. It does not present new squirrel experiments, a comprehensive AI benchmark implementation, or a theorem. Its contribution is instead fourfold. First, it synthesizes the squirrel literature around a single joint problem rather than as separate stories about agility, memory, or sociality. Second, it makes the inferential ladder explicit so the argument does not slide from behavior to engineering by rhetoric alone. Third, it offers a sharper formalization than a generic state-space sketch by representing latent dynamics, episodic memory, observer beliefs, option-level actions, and delayed verifiers in one model. Fourth, it turns the comparison into falsifiable AI hypotheses and benchmark families.

We argue for three primary hypotheses. H1 concerns fast local feedback and predictive compensation under hidden dynamics. H2 concerns structured episodic memory organized for future control rather than archival recall. H3 concerns the placement of verifiers and observer models inside the action-memory loop. A role-differentiated institutional extension is retained only as a downstream conjecture. That distinction matters. The paper's strongest warrant lies in the coupled control-memory-observability case within one organism; its weakest warrant lies where the argument jumps to macro-organizational AI design. The revision therefore keeps the latter bounded, explicit, and benchmarkable.

\section{Comparative method, scope, and admissible inference}

The method is a narrative comparative synthesis. That genre is legitimate only if its inferential rules are explicit. We therefore impose three admissibility conditions. First, computational invariance: a biological observation is relevant only if it instantiates a substrate-independent problem such as partial observability, delayed feedback, interference, or observer-dependent policy. Second, minimal inference first: each observation is mapped first to the weakest computational claim it supports; stronger architectural proposals are labeled as hypotheses rather than findings. Third, falsifiable consequence: every design claim must imply at least one benchmark distinction, ablation, or failure mode that could prove it wrong.

Squirrels are selected not because they are the only animals with spatial memory or agile control, but because the same comparative system exposes uncertain locomotion, distributed cache memory, and competitive observation in a single ecological loop. That conjunction is unusually useful for AI because it forces a joint problem formulation. A comparison centered on dexterity alone or memory alone would be easier to make, but scientifically weaker for the question at hand.

Throughout, ``squirrel'' refers to a family of studies on fox squirrels (\emph{Sciurus niger}), eastern gray squirrels (\emph{Sciurus carolinensis}), and, in Macdonald's field-memory comparison, Eurasian red squirrels (\emph{Sciurus vulgaris}). We infer a shared computational pattern only at the level of demand, not mechanism. Claims about locomotor adaptation derive from fox squirrels [1]. Claims about self-cache recovery are strongest in gray squirrels [2], [3]. Chunking-like cache organization and value-calibrated cache effort are fox-squirrel results [4], [5]. Audience effects are established in gray squirrels [6], with the broader socioeconomic framing provided by later review work [7]. Neuroanatomical anchors are species-specific and are treated as such later in the paper [8], [9].

\begin{figure}[H]
\centering
\includegraphics[width=0.75\textwidth]{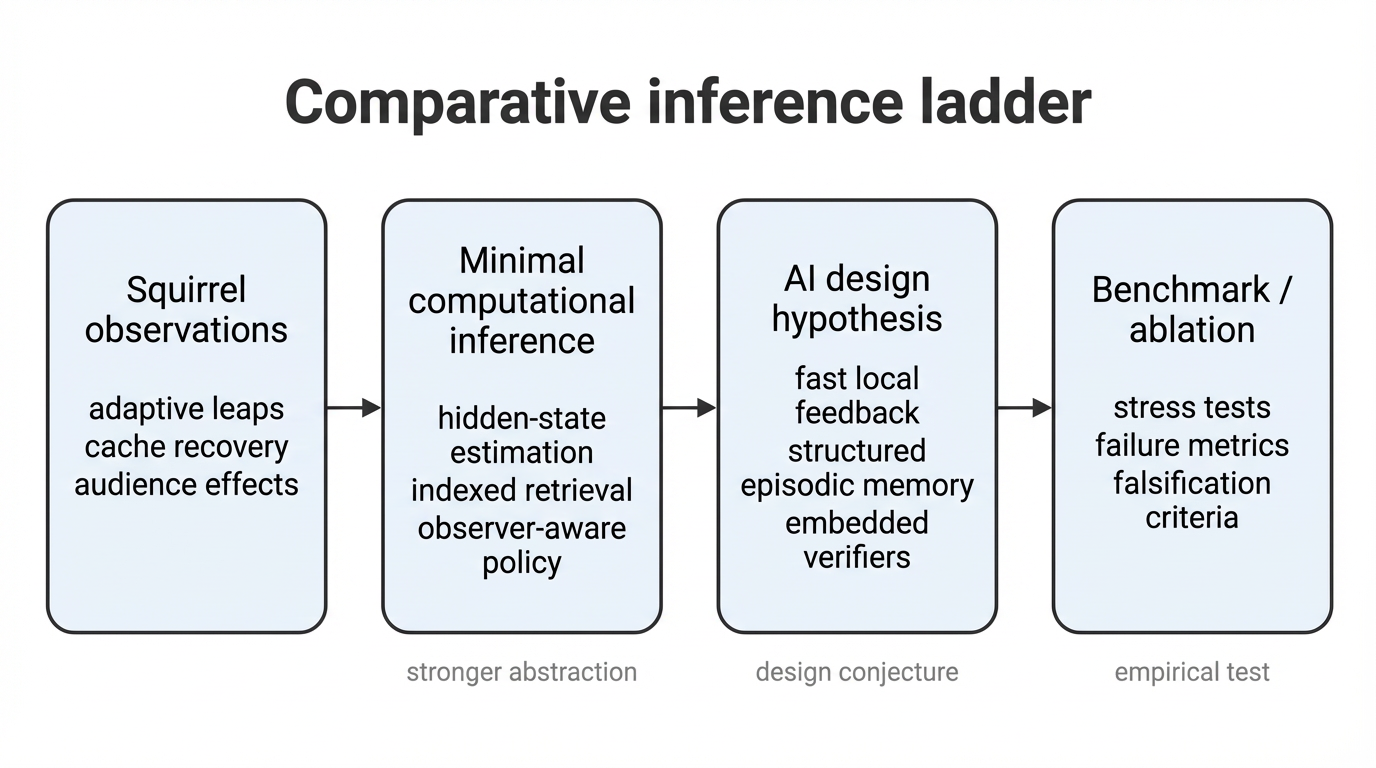}
\caption{Comparative inference ladder. Each upward step from observation to engineering claim increases abstraction and therefore strengthens the need for explicit benchmarking and ablation.}
\label{fig:inference-ladder}
\end{figure}

Figure~\ref{fig:inference-ladder} summarizes the inferential ladder that governs the manuscript, and Table~\ref{tab:evidence-map} records where empirical warrant is strong, moderate, or intentionally speculative. This discipline changes the argumentative burden. The question is not whether squirrel behavior literally implements a particular AI architecture. The question is whether the same joint problem, once made explicit, motivates a tractable program of AI benchmarks and ablations.

\begin{table}[H]
\caption{Evidence map. Confidence refers to the warrant for the minimal computational inference, not to the stronger AI hypothesis.}
\label{tab:evidence-map}
\footnotesize
\begin{tabularx}{\textwidth}{@{}L{0.23\textwidth}L{0.22\textwidth}L{0.25\textwidth}L{0.18\textwidth}@{}}
\toprule
\textbf{Biological observation} & \textbf{Minimal computational inference} & \textbf{AI hypothesis to test} & \textbf{Confidence / scope} \\
\midrule
Adaptive launch-point selection and landing recovery on compliant branches (fox squirrel) [1] & Online hidden-dynamics estimation and rapid error correction & Fast local feedback plus predictive compensation improves robustness under latent dynamics shifts & High for behavior; architecture still hypothetical \\ \\

Own-cache recovery and landmark-guided search after delay (gray squirrels; red-squirrel comparison in field study) [2], [3] & One-shot spatial memory, indexed retrieval, cue integration & Structured episodic memory reduces retrieval latency and cue-conflict failure & High for memory competence; moderate for specific indexing mechanism \\ \\

Spatial chunking by nut species under specific foraging regime (fox squirrel) [4] & Retrieval can benefit from clustered organization or compression & Typed or clustered memory indices outperform flat archives at equal storage budget & Medium; context-dependent behavioral result \\ \\

Value- and scarcity-calibrated cache effort (fox squirrel) [5] & Storage policy is coupled to expected future utility and action cost & Memory write and retrieval policies should be cost-sensitive rather than unconditional & Medium-high \\ \\

Audience-sensitive cache spacing and body orientation (gray squirrel) [6], [7] & Visible action carries information; observer model affects policy & Observer-aware policies and leakage penalties improve robustness under strategic observation & High for behavior; moderate for AI analogue \\
\bottomrule
\end{tabularx}
\end{table}

\section{Biological evidence for the coupled problem}

Three strands of evidence matter: arboreal locomotion under hidden dynamics, scatter-hoarding as delayed memory for future action, and audience-sensitive caching under theft risk. The argument does not depend on claiming identical mechanism across species. It depends on the conjunction of demands.

\subsection{Arboreal locomotion under latent support dynamics}

Free-ranging fox squirrels do not behave as open-loop ballistic scripts. In controlled field experiments with simulated branches, they adjusted launch location to a trade-off between gap distance and support compliance, improved over repeated exposure to unfamiliar flexible supports, and used agile landing repertoires to recover from error [1]. The ecological challenge is not merely force production; it is decision making under hidden mechanics. Compliance is partly latent at launch, sensory information arrives with delay, and waiting too long can be as costly as acting too early.

The minimal computational inference is online estimation and rapid correction under partial observability. The behavior is consistent with optimal-feedback and internal-model perspectives in motor control [10], [11], but the evidence does not warrant a claim that squirrels implement any particular controller. What it does warrant is more specific and more useful for AI: a competent system must combine predictive compensation with local stabilizing loops when dynamics are uncertain and time to intervention is short.

That point matters because many agentic failures in robotics and software look less like ordinary classification error than like poor recovery: state drift, brittle plan execution, overcommitment to stale beliefs, or inability to revise after a local perturbation. The squirrel case is informative precisely because it connects fast embodied error correction to learned adaptation across trials, rather than treating them as separate modules.

\subsection{Scatter-hoarding as memory for future control}

Scatter-hoarding supplies the memory side of the comparison. Gray squirrels recover their own buried nuts more accurately than nearby nuts buried by others, showing that retrieval is not reducible to opportunistic odor following alone [2]. In field conditions, squirrel spatial memory remains sufficiently precise to support targeted search after days-to-weeks delays, and natural landmarks can guide retrieval [3]. These are not passive recognition tasks; they are one-shot encodings that must support later action.

Fox-squirrel work further shows that memory organization is policy-relevant. Under specific foraging regimes, caches are spatially organized by nut species in a manner consistent with chunking-like mnemonic structure [4]. Assessment behaviors and cache effort scale with value and scarcity [5]. The squirrel is therefore not merely storing locations. It is allocating effort across present consumption, future recovery, concealment, and travel cost.

The minimal computational inference here is memory for future control. That interpretation is also consistent with broader work on episodic memory mechanisms that links encoding, retrieval, and later behavior through spatiotemporal trajectories rather than passive archival storage [24]. Useful memory must index episodes, manage interference, support fast retrieval under cue conflict, and remain coupled to expected future value. The predictive-map account of hippocampal computation is relevant because it treats memory as structure for future occupancy and planning rather than archival storage [12]. Again, the behavioral literature does not prove a specific indexing scheme. It does show that delayed competent action depends on memory organization, not on raw storage volume alone.

Scatter-hoarding is therefore best understood as an economic control problem. Cache placement changes future path length and theft risk; memory errors waste both energy and opportunity. A design lesson for AI follows immediately: memory modules should be evaluated not as passive databases but as control resources whose structure affects later action quality, latency, and repair cost.

\subsection{Audience-sensitive caching and strategic observability}

Social observability makes the case stronger. Gray squirrels space caches farther apart and preferentially orient their bodies away from observing conspecifics when caching, consistent with pilferage-avoidance strategies [6]. Reviews of wild food hoarding emphasize that caching decisions are shaped by competition, theft risk, predation, and environmental structure [7]. A visible action is therefore also an information release.

The minimal computational inference is observer-aware policy modulation. The future value of a cache depends not only on whether the squirrel can later find it, but also on whether another squirrel can infer it from the current act of storage. For AI, the relevant parallel is not anthropomorphic deception. It is the ordinary fact that tool use, memory writes, and visible outputs can leak task-relevant information to competitors, attackers, or unauthorized observers. Strategic observability is not an edge case; it is part of routine competent action.

Taken together, locomotion, scatter-hoarding, and audience effects form a tighter comparative case than biological analogies centered on a single salient faculty. The same ecological loop requires rapid correction, delayed retrieval, valuation, and observer-sensitive policy. That conjunction, not any one behavior alone, is the paper's central empirical foundation.

\section{Minimal formalization: hierarchical partially observed control with memory and verification}

The formal layer should therefore make explicit the variables that are distinctive to this joint problem. A generic state-space model is not enough. We need at least an embodied control state, a latent environmental variable, an episodic memory state with update and retrieval operators, an observer-belief state, option-level actions, and delayed verifier signals.

For convenience, we refer to this coupled perspective as SCRAT (Stochastic Control with Retrieval and Auditable Trajectories). The framing is naturally read as a stochastic control problem in which a stateful agent acts under partial observability, retrieves control-relevant memory, and updates policy under noisy delayed feedback [25].

Here $x_t$ denotes embodied plant state such as pose, velocity, or actuator state; $z_t$ denotes latent environmental dynamics and cue reliability, such as support compliance, friction, or landmark stability; $m_t$ denotes structured episodic memory; $b_t$ denotes an estimate of what relevant observers or adversaries can infer; and $e_t$ denotes task, resource, and permission state. The state decomposition can therefore be written as
\begin{align}
 s_t &= (x_t, z_t, m_t, b_t, e_t), &
 \beta_t &= p(s_t \mid o_{1:t}, a_{1:t-1}).
 \label{eq:state}
\end{align}
Here $o_t$ denotes observation. The belief state $\beta_t$ summarizes what the agent currently infers under partial observability.

Control unfolds at two levels. An option $w_t$ is a temporally extended macro-action such as launch, stabilize, cache, retrieve, or conceal [14]. The planner chooses options; a local controller executes primitive actions conditioned on current belief and retrieved memory; and the retrieval policy determines which memories become control-relevant. This is the level at which the squirrel comparison most naturally connects to hierarchical control, temporal abstraction, and integrated model-based architectures [13]--[16]. A minimal control loop can therefore be written as
\begin{align}
 w_t &\sim \Pi(\cdot \mid \beta_t), &
 q_t &= Q(\beta_t, w_t), &
 r_t &= \mathcal{R}(m_t, q_t), &
 a_t &\sim \pi(\cdot \mid \beta_t, r_t, w_t).
 \label{eq:control}
\end{align}
Here $q_t$ is the retrieval query induced by current belief and option, and $r_t$ is the retrieved control-relevant memory.

The coupled update makes two variables explicit that the original draft only implied. First, memory is dynamic: it is updated by observation, action, and later verification. Second, verification can be delayed. In ecological settings, outcomes such as successful landing, later cache recovery, or pilferage are noisy, delayed, and partial checks. In AI systems, their analogues are precondition tests, permissions, runtime monitors, provenance constraints, and postcondition evaluators [17], [18], [22]. None is an oracle. All can be incomplete, noisy, or gameable. But without some checker pathway, delayed action cannot be made legible to either the agent or its overseers. The coupled update can therefore be expressed as
\begin{align}
 \beta_{t+1} &= \mathcal{B}(\beta_t, o_{t+1}, a_t), \nonumber\\
 m_{t+1} &= U(m_t, o_{t+1}, a_t, v_{t+\Delta}), \label{eq:update}\\
 b_{t+1} &= \mathcal{O}(b_t, a_t, o_{t+1}), \nonumber\\
 v_{t+\Delta} &= V(\tau_{t:t+\Delta}, \eta). \nonumber
\end{align}
Here $\tau_{t:t+\Delta}$ denotes the executed trace segment later exposed to checking, and $\eta$ denotes verifier specification or tolerance parameters.

The objective function makes the trade-offs explicit:
\begin{align}
 \min_{\Pi,\pi,Q,\mathcal{R},U,\mathcal{O}} \; \mathbb{E}\Bigg[\sum_{t=0}^{T-1} \Big(c_{\text{task},t} + \lambda_{\tau} c_{\text{latency},t} + \lambda_{L} c_{\text{leak},t} + \lambda_{R} c_{\text{repair},t}\Big)\Bigg]
 \label{eq:objective}
\end{align}
subject to
\begin{align}
 \Pr(v_T = 1) &\ge 1-\delta, &
 \mathbb{E}\!\left[\sum_t \kappa_t\right] &\le B.
 \label{eq:constraints}
\end{align}
Here $c_{\text{task},t}$, $c_{\text{latency},t}$, $c_{\text{leak},t}$, and $c_{\text{repair},t}$ denote task, latency, leakage, and repair costs respectively; $\kappa_t$ denotes per-step compute burden; and $B$ denotes a total compute budget.
Competence is not only task success. It also includes latency, leakage, and repair cost after late failure. This is precisely where squirrel ecology is instructive. A cache recovered too slowly or a leap corrected too late is still a bad policy; an action that reveals too much can be locally successful and globally unwise. The formal point is not that squirrels instantiate this exact model. It is that any architecture claiming competence on the same joint problem must represent or absorb these burdens somewhere, whether explicitly through modules or implicitly through a monolithic policy.

On this view, a world model should be interpreted narrowly: predictive, action-conditioned structure with operational value for control. The squirrel case does not require a single monolithic simulator. It requires enough predictive structure to compensate for hidden dynamics, guide retrieval under uncertainty, and anticipate the informational consequences of visible action. In that operational sense, the practical distinction between an LLM and a world model is narrower than nomenclature often suggests: once a stateful sequential model updates internal state from observation and uses that state for action-conditioned prediction, it is already functioning as a world model for the task at hand [25]. That places the proposal at the intersection of optimal feedback control [10], partially observed planning [13], temporal abstraction [14], Dyna-style integration of learning, planning, and reacting [15], episodic-memory agents [16], and world-model architectures [19], [20]. The proposal is also adjacent to earlier work on algorithmic information, adaptive Levin search, speed priors, and incremental self-improvement in general problem solving [26]-[30]. The contribution here is not a new theorem about those frameworks. It is a joint evaluation target.

\begin{figure}[H]
\centering
\includegraphics[width=0.97\textwidth]{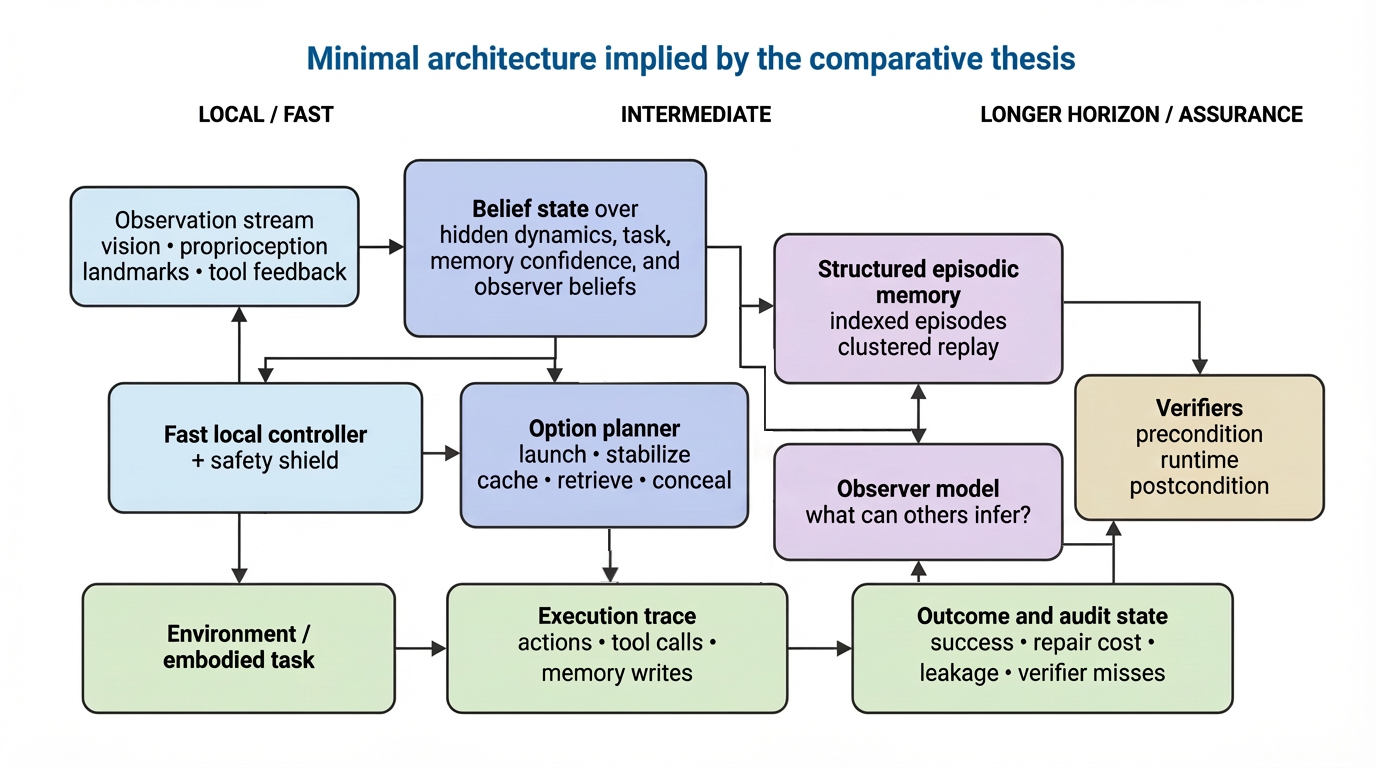}
\caption{Minimal architecture implied by the comparative thesis. The proposal is not a claim of literal squirrel mechanism; it is an engineering decomposition that makes H1-H3 and C1 benchmarkable.}
\label{fig:architecture}
\end{figure}

\section{Testable design hypotheses for agentic AI}

The formalization earns its place only if it yields benchmark-relevant distinctions. We therefore state the main claims as hypotheses rather than findings. The first three are the paper's primary claims. The fourth is a downstream conjecture whose warrant is deliberately weaker.

\subsection{H1: Fast local feedback plus predictive compensation improves robustness under latent dynamics shifts}

Architectures that place a short-horizon controller near execution should degrade less under unseen mechanics than architectures that rely primarily on open-loop planning or delayed correction. In control terms, the relevant advantage comes from coupling global planning to local closed-loop correction rather than relying on open-loop execution alone. The expected gain is largest when the hidden variable is locally consequential and quickly changing, precisely the situation in compliant-branch locomotion and many embodied manipulation tasks. The relevant failure signature is not merely lower final success. It is slower recovery, more interventions, and larger repair cost after small perturbations.

This hypothesis is compatible with control-theoretic work on minimal intervention and internal models [10], [11], and with Dyna and world-model architectures that plan through predictive dynamics [15], [19], [20]. The squirrel comparison adds a sharper benchmark criterion: the controller must remain competent when dynamics are partly latent at action time and when the window for local correction is brief.

\subsection{H2: Structured episodic memory improves delayed retrieval under cue conflict and memory load}

Memory should be engineered as an indexed control resource rather than an undifferentiated archive. In squirrel-like tasks, what matters is not only whether a past episode was stored, but whether the right episode can be retrieved quickly when landmarks drift, cues conflict, or thousands of similar episodes compete. The expected gain from structure is lower cross-item confusion, slower growth of retrieval latency with memory load, and better graceful degradation under interference.

Chunking-like cache organization [4] and value-sensitive cache effort [5] suggest that storage, indexing, and later action are co-designed. AI analogues include retrieval policies, clustered replay, typed memory slots, and experience reuse mechanisms that privilege control-relevant episodes [15], [16]. The comparative claim is not that squirrels prove a particular indexing scheme. It is that a flat archive is the wrong baseline for delayed action under interference.

\subsection{H3: Verifiers and observer models should sit inside the action-memory loop}

If an agent's actions can leak information and if success is only partially checkable, verification cannot sit only at the end of the pipeline. Precondition checks should screen obviously impermissible actions; runtime monitors should detect deviation; postcondition checks should evaluate delayed success; provenance traces should preserve the basis of memory-mediated actions; and observer models should estimate what an adversary or unauthorized viewer can infer from visible behavior.

This hypothesis is easy to overstate. Verifiers are not guaranteed safeguards. They can be incomplete, produce false positives or false negatives, and induce proxy optimization in which the agent learns to satisfy the checker while missing the real objective [17], [18]. The point is therefore not ``add one checker and solve alignment.'' It is that verifiability must be treated as an architectural burden, with heterogeneous checks and explicit failure accounting, rather than as an afterthought.

The squirrel comparison sharpens one additional implication: privacy and leakage are control variables, not merely policy constraints. Audience-sensitive caching shows that an action can be locally successful yet globally unwise because it reveals too much [6], [7]. For AI systems, the corresponding metric is not just task reward, but the joint profile of reward, leakage, verifier miss rate, and repair cost.

\subsection{C1: Role-differentiated proposer/executor/checker/adversary systems are a downstream conjecture, not a direct biological inference}

The squirrel evidence does not by itself prove that AI institutions should be decomposed into specialized roles. What it does suggest is that competence often depends on heterogeneous timescales, asymmetric information access, and distinct verification burdens. That observation makes role differentiation a plausible systems hypothesis: a proposer may search broadly, an executor may act conservatively, a checker may enforce specifications, and an adversary may surface blind spots.

The conjecture has antecedents in the society-of-mind tradition [23], adversarial oversight schemes such as debate [21], and broader proposals for verifiable claims and external scrutiny in AI development [22]. Its warrant here is nonetheless weaker than H1-H3. It should be judged only by benchmark results showing reduced correlated error or lower silent-failure rates relative to monolithic or homogeneously incentivized systems.

\section{Benchmark agenda and evaluation criteria}

The empirical program should couple, rather than isolate, the burdens highlighted above. A strong benchmark suite would contain at least four families. Family A would stress hidden-dynamics control in an arboreal or branch-like environment with variable compliance, friction, delay, and observation noise. Family B would stress cache-like episodic retrieval at scale through thousands of one-shot storage events followed by delayed queries under landmark drift and cue conflict. Family C would stress observer-aware action under strategic observation, where visible behavior or memory traces can be exploited by a competitor. Family D would test the downstream conjecture about role differentiation under imperfect checker coverage and hidden constraints.

Across families, the decisive metrics are not static accuracy alone. They are success rate, time to successful completion, retrieval latency, repair cost after perturbation, silent-failure frequency, information leakage to an adversary, verifier false-positive and false-negative rates, and compute overhead. Because H1-H3 concern time as well as correctness, latency distributions should be reported alongside final success.

Each family should include clean ablations: remove fast feedback; remove predictive compensation; flatten memory into an archive; disable observer models; delay all checking to the end; or collapse differentiated roles into one agent. Report confidence intervals, seed counts, effect sizes, compute budgets, and representative failures. If the proposed decomposition is real, its benefits should survive across random seeds and modest task reparameterizations rather than appearing only in a single tuned environment.

\begin{table}[H]
\caption{Benchmark families, manipulated variables, metrics, and critical ablations.}
\label{tab:benchmark-families}
\footnotesize
\begin{tabularx}{\textwidth}{@{}L{0.17\textwidth}L{0.18\textwidth}L{0.18\textwidth}L{0.18\textwidth}L{0.17\textwidth}@{}}
\toprule
\textbf{Benchmark family} & \textbf{Manipulations} & \textbf{Components stressed} & \textbf{Core metrics} & \textbf{Critical ablations} \\
\midrule
A. Hidden-dynamics control & Support compliance, friction, observation delay, perturbation timing & Fast local feedback, predictive model, latent-state adaptation & Success rate, time-to-stabilization, intervention count, repair cost & Remove feedback; remove latent-dynamics adapter \\ \\

B. Cache-like episodic retrieval at scale & Number of stored events, delay length, landmark drift, cue conflict & Memory indexing, retrieval policy, compression and replay & Recall precision, retrieval latency, cross-item confusion, degradation curve & Flat archive; no clustered or typed indices \\ \\

C. Observer-aware action under strategic observation & Observer visibility, adversary sophistication, hidden constraints, spoofed cues & Observer model, privacy gate, verifier placement & Reward, leakage, verifier miss rate, post-hoc correction cost & No observer model; end-only checking \\ \\

D. Role-differentiated verification pipeline & Checker coverage, shared vs. separate memory, role incentives, adversarial pressure & Proposer, executor, checker, adversary split & Silent-failure frequency, correlated error, compute overhead, repair cost & Single agent; shared incentives across all roles \\
\bottomrule
\end{tabularx}
\end{table}

\subsection{Preliminary systems evidence for structured memory at project scale}

A first instantiation of Family B already exists in Chiron, a software-delivery system described by the present authors in a companion engineering benchmark [31]. For each project, the system clones the repository, ingests the existing project documentation, analyzes the codebase file by file, partitions source and document artifacts into semantically typed chunks, assigns metadata indicating module membership, business-logic role, inter-artifact connectivity, and project-level relevance, and then builds a graph-structured memory over those relations. That memory is queried dynamically during document generation, task execution, testing, and validation, so retrieval is conditioned on the current workflow stage and task context rather than fixed at ingestion time. The comparison is informative for the present paper because it contrasts an isolated-agent baseline, in which agents act with only local context and no persistent memory or integrated review, against a memory-augmented review-integrated configuration in which repositories and documentation are chunked, relevance-labeled, business-logic-tagged, graph-linked, and retrieved later under task-specific queries [31]. Retrieval in that system is both delayed and context-sensitive: information written into memory at ingestion is retrieved later under changing documentation, implementation, testing, and validation states, and the labeling, relevance scoring, and graph relations are used to reduce cue conflict by surfacing the right episodes for the current task.

\begin{table}[H]
\caption{Preliminary structured-memory evidence from a companion software-delivery benchmark by the present authors [31]. Durations are measured in weeks. Issue load is reported in staged form as isolated-agent baseline $\rightarrow$ mature configuration before downstream validation $\rightarrow$ mature configuration at downstream validation; the middle value is derived from approximate issue counts recorded before downstream validation.}
\label{tab:preliminary-memory}
\footnotesize
\begin{tabularx}{\textwidth}{@{}L{0.14\textwidth}L{0.12\textwidth}L{0.09\textwidth}L{0.14\textwidth}L{0.15\textwidth}L{0.24\textwidth}@{}}
\toprule
\textbf{Program} & \textbf{Approx. scale} & \textbf{Chunks} & \textbf{Duration} & \textbf{First-release coverage} & \textbf{Issue load / 100 tasks} \\
\midrule
Bank application & $\sim$30k LOC & $\sim$600 & 7.4 $\rightarrow$ 2.4 & 55\% $\rightarrow$ 90\% & 10.00 $\rightarrow$ 5.00 $\rightarrow$ 2.50 \\
ACAS & $\sim$400k LOC & $\sim$8000 & 16.5 $\rightarrow$ 4.9 & 50\% $\rightarrow$ 90\% & 8.44 $\rightarrow$ 4.18 $\rightarrow$ 2.00 \\
Mortgage application & $\sim$30k LOC & $\sim$400 & 4.7 $\rightarrow$ 2.0 & 70\% $\rightarrow$ 95\% & 8.18 $\rightarrow$ 4.18 $\rightarrow$ 2.18 \\
Portfolio & -- & -- & 28.6 $\rightarrow$ 9.3 & 52.6\% $\rightarrow$ 90.5\% & 8.63 $\rightarrow$ 4.29 $\rightarrow$ 2.09 \\
\bottomrule
\end{tabularx}
\end{table}

\begin{figure}[H]
\centering
\includegraphics[width=\textwidth]{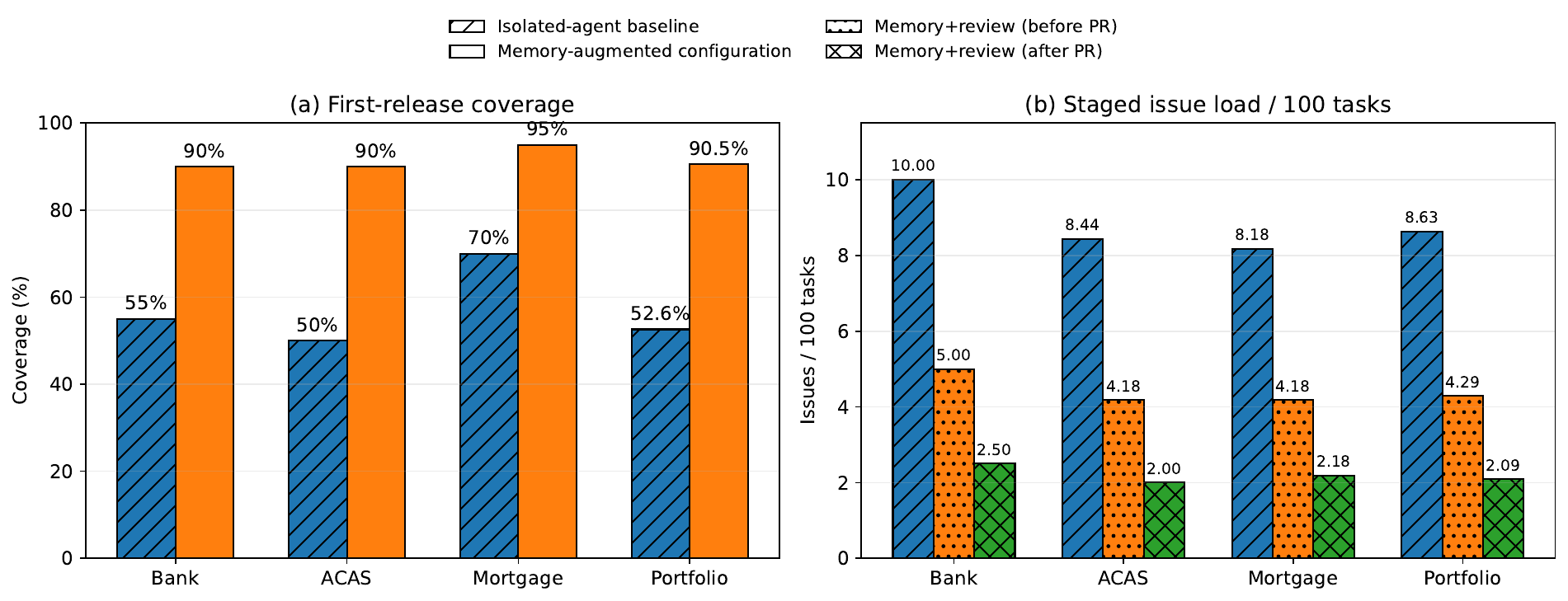}
\caption{First-release coverage and staged issue load in the isolated-agent baseline and the memory-augmented review-integrated configuration from the companion software-delivery benchmark [31]. The middle bar in the issue-load panel reports issue burden before the pull-request boundary, so it isolates the structured-memory effect before the additional defect containment introduced by review; the rightmost bar reports the remaining issue burden at downstream validation after review.}
\label{fig:preliminary-memory}
\end{figure}

Across the three programs, summed project duration falls from 28.6 to 9.3 weeks, a 3.08$\times$ speedup. Task-weighted first-release coverage rises from 52.6\% to 90.5\%, and validation-stage issue load falls from 8.63 to 2.09 issues per 100 tasks [31].

Because the companion benchmark also records approximate issue counts present before the pull-request boundary in the memory-augmented review-integrated configuration, the quality comparison can be read in staged form rather than only through downstream validation. That distinction is included for a specific reason relevant to the present paper: the pre-PR value is the closest available estimate of the structured-memory contribution itself, because it measures issue burden after retrieval-guided documentation, implementation, testing, and validation, but before additional defect containment from human review and the specialized review agent. The post-PR value then shows the further reduction once that review boundary is added. At project level, the memory-augmented configuration carries approximately 5.00, 4.18, and 4.18 issues per 100 tasks before downstream validation in Bank, ACAS, and Mortgage, compared with downstream loads of 10.00, 8.44, and 8.18 in the isolated-agent baseline; after the review boundary, the corresponding validation-stage loads are 2.50, 2.00, and 2.18 [31]. Portfolio-wide, the same sequence is 8.63 issues per 100 tasks in the isolated-agent baseline, 4.29 before downstream validation in the memory-augmented configuration, and 2.09 after the review boundary.

In operational terms, the mature workflow introduces an explicit review boundary rather than a generic repository hook: once an agent completes a task, it opens a pull request, a human engineer supported by a specialized review agent evaluates the change, and only then is the task merged and moved to done. That matters for the interpretation of Family B because the memory layer is used before documentation, during execution, and again during testing and validation, while the review stage provides an additional opportunity to intercept errors before downstream validation. The result is aligned with H2: project-scale indexing, retrieval policy, graph-structured memory, and stage-conditioned access to stored context coexist with faster completion, higher first-release coverage, and lower issue burden both before and after downstream validation.

\section{Threats to validity and limits}

Nothing in squirrel success establishes human acceptability. Ecological fitness is not a human objective, and strategic concealment in animals is not a normative license for deception in deployed AI. The comparison concerns architectural competence under coupled control-memory-verification burdens; moral sufficiency belongs to an additional layer of design and governance.

The evidence also pools across squirrel species and across behavioral domains. That is acceptable only at the level of shared computational demand. It is not a claim of species-uniform mechanism. Moreover, squirrel neurobiology offers bounded anchors rather than full circuit explanation: motor-cortex organization has been mapped in squirrels [8], and hippocampal morphometry in food-caching squirrels provides species-relevant memory anatomy [9], but circuit-level causal accounts remain sparse relative to mouse or primate work.

Verifier mismatch is a separate threat. Ecological outcomes are not designed software tests. A landing failure, a lost cache, or a pilfered nut is a noisy consequence, not a formal specification. Engineered verifiers therefore introduce a mismatch: they can underapproximate harm, overapproximate constraint violation, or be gamed by agents that optimize the checker instead of the real objective [17], [18], [22]. Any practical program built from this perspective must measure checker failures explicitly, not merely report final task reward.

The thesis would be weakened by at least three findings. First, monolithic architectures without explicit control-memory-verifier decomposition might match or exceed decomposed systems on joint benchmarks under hidden dynamics, heavy memory load, and strategic observation while maintaining equal or lower silent-failure rates. Second, structured episodic memory might provide no measurable advantage in retrieval latency, interference resistance, or graceful degradation once total parameter count and context length are controlled. Third, role differentiation might fail to reduce correlated error, or help only when the checker is unrealistically strong and non-gameable. These are real possibilities. The value of the squirrel comparison lies in making them testable.

\section{Discussion: from within-agent specialization to cautious role differentiation}

Within-agent specialization and institution-level specialization should not be conflated. The squirrel evidence strongly supports the former and only weakly motivates the latter. Still, the bridge is worth stating carefully. Systems become harder to govern when the same module plans, executes, remembers, and judges itself under one incentive. When information access and verification burdens differ, differentiated roles can create opportunities for scrutiny and repair.

That observation does not justify institutional exuberance. Diversity of voices does not guarantee reliability; heterogeneous agents can still share a blind spot, collude, or reproduce the same proxy error. The appropriate interpretation is narrower: if a task naturally separates broad search, cautious execution, specification checking, and adversarial challenge, then architectures that preserve that separation may be easier to evaluate and audit.

Seen this way, the squirrel comparison contributes a constraint rather than a grand theory. Role boundaries should be drawn by timescale, information access, and verification burden. Wherever those factors do not meaningfully differ, added plurality may simply add cost and latency. Where they do differ, monolithic agency may be the more fragile design.

\section{Conclusion}

The central claim is not that squirrels are miniature AI systems. It is that squirrel ecology sharpens a joint engineering problem that AI research still too often decomposes prematurely. Rapid control under hidden dynamics, memory for future options, and delayed partial verification are not separate extras; in many real tasks they are one problem.

The squirrel literature warrants a coupled control-memory-observability thesis, a minimal formal model, three primary AI hypotheses, and a downstream conjecture on role differentiation. One preliminary software-delivery example reported here suggests that the structured-memory component can already be instantiated at project scale, but the broader coupled thesis still stands or falls on benchmark evidence. That is the standard under which this perspective should be judged.

\end{document}